\documentclass[10pt,twocolumn,letterpaper]{article}

\pdfoutput=1

\usepackage[pagenumbers]{cvpr} 

\newcommand{\method}{SilentWAM\xspace}
\newcommand{\mtok}{\texttt{[M]}}

\usepackage{tikz}
\usetikzlibrary{positioning,fit,calc,arrows.meta,decorations.pathreplacing}

\usepackage{pgfplots}
\pgfplotsset{compat=1.18}
\usepgfplotslibrary{groupplots}
\definecolor{cwm}{HTML}{2A78D6}
\definecolor{cnowm}{HTML}{8A8986}
\definecolor{cd1}{HTML}{2A78D6}
\definecolor{cd2}{HTML}{EB6834}
\definecolor{cd3}{HTML}{1BAF7A}

\usepackage{tabularx}
\newcolumntype{Y}{>{\raggedright\arraybackslash}X}

\definecolor{cvprblue}{rgb}{0.21,0.49,0.74}
\usepackage[pagebackref,breaklinks,colorlinks,allcolors=cvprblue]{hyperref}

\def\confName{CVPR}
\def\confYear{2027}

\title{\method: Planning in the Masked State of a Diffusion World Model}

\author{%
Ziyu Zhao \quad Casey Sun \quad Zeyu Wang \quad Shicong Ma \quad Emily T. Zhang \quad Cheng Lu\\
Zoox\\
{\tt\small \{zzhao, casun, zewang, sma, ezhang, clu\}@zoox.com}%
}

\begin{document}
\twocolumn[{%
\renewcommand\twocolumn[1][]{#1}%
\maketitle
\begin{center}
  \centering
  \includegraphics[width=0.92\textwidth,trim={12pt 1pt 8pt 0pt},clip]{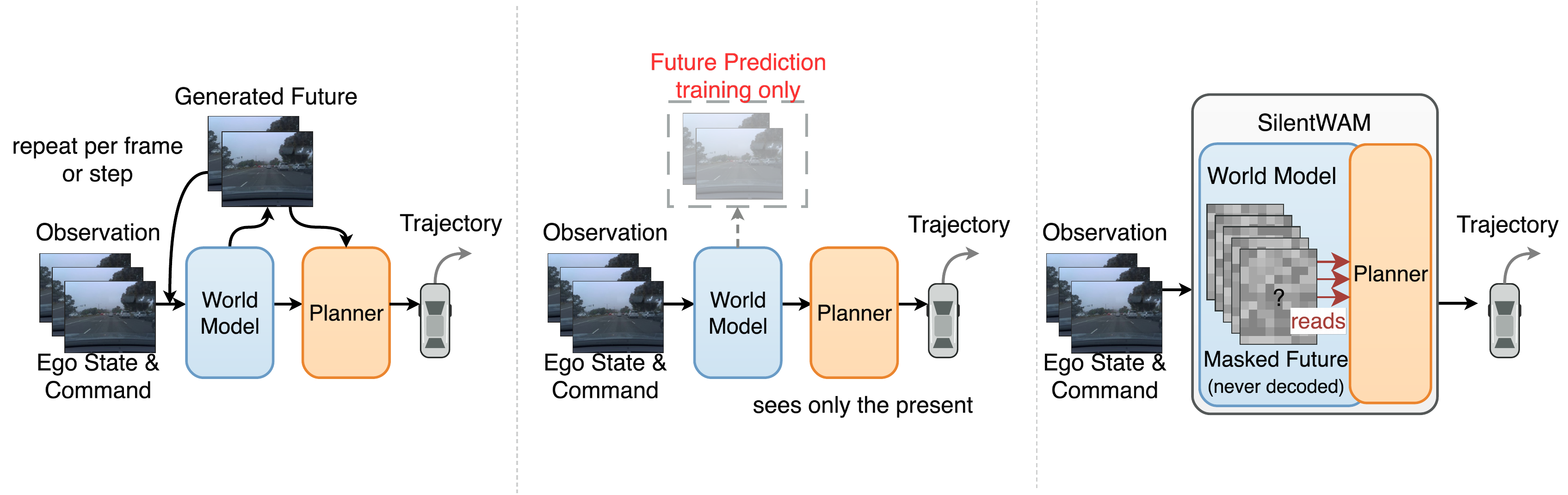}\\[-1pt]
  {\small
  \parbox[t]{0.3167\textwidth}{\centering (a) Imagine-then-act}%
  \parbox[t]{0.3167\textwidth}{\centering (b) Future co-training}%
  \parbox[t]{0.3167\textwidth}{\centering (c) \method (ours)}}
  \captionof{figure}{How a world-action model uses the predicted future at test time. (a) Future frames are generated before planning, requiring additional forward passes. (b) Future prediction is used only during training; explicit future predictions are omitted at inference. (c) \method: the planner reads the masked future in the same forward pass, without decoding it.}
  \label{fig:teaser}
\end{center}
\vspace{-3mm}
}]
\begin{abstract}
World-action models learn from future prediction, but accessing their future representations directly at test time can require costly frame generation.
Methods that generate future frames before planning incur additional forward passes, while future co-training methods omit explicit future predictions at inference.
We introduce \method, a planner that reads the hidden states of masked future tokens without decoding future frames.
We pretrain MMaDA-Parallel-M on driving video using frame-level masked world modeling, which predicts a masked suffix of future frames and explicitly includes inputs with all future frames masked.
Learned planning queries attend to these masked positions and the observed context, and a continuous head predicts the trajectory in the same forward pass.
With a single front camera, \method reaches 92.0 PDMS on NAVSIM v1 and 89.3 EPDMS on NAVSIM v2, with an end-to-end latency of 60.3\,ms.
\end{abstract}

\section{Introduction}
\label{sec:intro}

End-to-end driving models predict a trajectory directly from sensor input~\cite{hu2023uniad,jiang2023vad,liao2025diffusiondrive}.
Much of the difficulty is anticipating how the scene will develop, such as whether an adjacent car will cut in or the lead vehicle will brake.
World models learn such dynamics from unlabeled driving video by predicting future observations~\cite{hu2023gaia1,gao2024vista}, and world-action models (WAMs) train a single network to predict both future observations and actions~\cite{zhao2025pwm,li2026drivevlaw0,zhao2026simwam}.
For onboard deployment, the planner needs access to these predictions, and producing them must fit the inference-time budget.

Current WAMs handle these two requirements differently (\cref{fig:teaser}).
Imagine-then-act models~\cite{zhao2025pwm,zhang2025epona,yao2026discretewam} first generate future frames and then plan from them (\cref{fig:teaser}a). This gives the planner access to a predicted future, but requires additional generation steps.
Future co-training models~\cite{li2025law,li2026drivevlaw0,zhao2026simwam,yuan2026fastwam} predict the future only during training and discard it at test time (\cref{fig:teaser}b), allowing single-pass planning without explicit future predictions.
ForeDrive~\cite{wang2026foredrive} and RIFT~\cite{zhang2026rift} keep a predicted future at test time without generating frames, using a separately trained latent predictor and learned anticipation tokens, respectively.
We study whether a planner can reuse the world model's future representations without decoding frames or adding a separate future predictor.

Autoregressive and masked-diffusion world models expose future predictions in different ways.
The autoregressive world models considered here~\cite{hu2023gaia1,zhao2025pwm,zhang2025epona} predict future frames sequentially, conditioning each prediction on preceding frames. Accessing their multi-frame predictions therefore requires a rollout.
Masked diffusion models~\cite{nie2025llada,yang2025mmada} are trained differently: arbitrary subsets of tokens are masked and predicted from the visible ones. We explicitly train on the case in which all future frames are masked.
The hidden state at each masked future position is trained to support prediction of the corresponding token~\cite{ou2025absorbing}, for example whether a neighboring car will occupy that image region a second later.
We refer to these hidden states as the \emph{masked future state}.
A planner that attends to them can use the predicted future within the same forward pass that produces the trajectory, while the future frames themselves are never decoded (\cref{fig:teaser}c), hence the name \method.

Masked diffusion has been used in driving before~\cite{li2025reflectdrive,ma2025dvlmad,xu2025wamdiff,zhang2026fastddrive}, but as a decoder for actions and text, on vision-language backbones that read the camera image without being able to generate it (\cref{sec:related}).
Our approach uses an image-generative backbone to learn representations at masked future positions.
We therefore build \method on MMaDA-Parallel-M~\cite{tian2025mmadaparallel}, an 8B masked diffusion model that generates text and images as tokens of one vocabulary, with bidirectional attention over the whole sequence.
To our knowledge, \method is the first to plan from the masked future state of a unified image-text masked diffusion model, reading the predicted future in one pass instead of decoding it.
We pretrain it on driving video with frame-level masked world modeling, which masks a suffix of future frames and predicts it from the observed ones (\cref{sec:wm}). Its fully masked case uses the same future-token configuration as planning.
For planning, learned queries attend to the masked future state together with the current frame, the ego state and the navigation command, and a continuous head regresses the waypoints (\cref{sec:plan}), followed by a short post-training stage with simulator rewards (\cref{sec:rl}).

We evaluate \method on NAVSIM v1, NAVSIM v2, and nuScenes using only the front camera (\cref{sec:main}).

In summary, our contributions are as follows.
\begin{itemize}
\item We show that the masked future state of a masked diffusion world model can serve as the input of a planner, which then uses the predicted future in one forward pass without generating frames or adding a separate predictor.
\item We introduce frame-level masked world modeling, a pretraining objective that explicitly trains on fully masked futures and predicts multiple future frames from the observed context.
\item We report 92.0 PDMS on NAVSIM v1 with one camera at 60.3\,ms per plan. In ablations, retaining the masked future lowers the nuScenes collision rate from 0.18\% to 0.06\%. Decoding future frames adds generation steps for only a marginal improvement in NAVSIM PDMS.
\end{itemize}

\section{Related Work}
\label{sec:related}

\paragraph{End-to-end planning.}
Modular end-to-end planners pass the scene through intermediate representations such as BEV maps, tracked agents and motion forecasts before planning~\cite{hu2022stp3,hu2023uniad,jiang2023vad}, and TransFuser~\cite{chitta2023transfuser} fuses camera and LiDAR features.
Generative planners produce multimodal trajectories with diffusion or flow matching~\cite{liao2025diffusiondrive,xing2025goalflow}, and vision-language-action models plan with a pretrained multimodal backbone~\cite{zhou2025autovla,li2026drivevlaw0}.
Planners that receive the ego state can reach low open-loop errors from the vehicle's own motion alone~\cite{li2024egostatus}, a form of the copycat problem in imitation from observation histories~\cite{wen2020copycat}.
Closed-loop reward optimization has recently been added on top of imitation~\cite{zou2025diffusiondrivev2,zhou2025autovla}.
\method uses a single front camera, reads the plan from learned queries with a continuous head, and adapts the reward optimization of DiffusionDriveV2~\cite{zou2025diffusiondrivev2} to this deterministic head for post-training.

\paragraph{World models for planning.}
Driving world models were first built to generate scenes~\cite{hu2023gaia1,wang2024drivedreamer,gao2024vista} or to forecast occupancy~\cite{zheng2024occworld,yang2025driveoccworld}.
Several planners forecast compact BEV or latent states with a small module and use them at test time: WoTE~\cite{li2025wote} rolls out future BEV states to score candidate trajectories, SeerDrive~\cite{zhang2025seerdrive} feeds the predicted BEV back to the planner, and World4Drive~\cite{zheng2025world4drive} scores latent futures of candidate intentions.
These states are supervised with BEV semantic maps or with pseudo-labels from vision foundation models, so the resulting future representations are shaped by the chosen labels or teacher features.
World-action models instead learn the dynamics from the observations themselves, and use them to plan in three ways.
Imagine-then-act models generate the future before acting: PWM~\cite{zhao2025pwm} forecasts frames one at a time with a block-causal mask and then predicts the trajectory, Epona~\cite{zhang2025epona} rolls out an autoregressive diffusion model, and Discrete-WAM~\cite{yao2026discretewam} generates frames frame by frame while editing action tokens in parallel.
Future co-training models predict the future only as a training signal and drop it at test time, predicting latent features~\cite{li2025law} or images~\cite{li2026drivevlaw0}, or co-training a video expert whose future tokens are hidden from the action expert~\cite{zhao2026simwam,yuan2026fastwam}.
A third group keeps a predicted future at test time without a full rollout: ForeDrive~\cite{wang2026foredrive} predicts latent futures with a separate predictor and injects them into the planner, and RIFT~\cite{zhang2026rift} fills the future positions with learned anticipation tokens in one pass.
\method reuses the world model's masked future positions, without a separate future predictor or additional anticipation tokens.

\paragraph{Masked diffusion models.}
Masked diffusion models corrupt discrete data by replacing tokens with a mask and learn to recover them~\cite{austin2021d3pm,sahoo2024mdlm,shi2024md4}; MaskGIT~\cite{chang2022maskgit} applies the idea to image tokens.
At scale, LLaDA~\cite{nie2025llada} is competitive with autoregressive language models of the same size, and LLaDA-V~\cite{you2025lladav} adds a continuous vision encoder for image understanding.
Unified models also generate images as discrete tokens: Show-o~\cite{xie2025showo} generates text autoregressively and predicts masked tokens within each image, while MMaDA~\cite{yang2025mmada} and MMaDA-Parallel~\cite{tian2025mmadaparallel} apply masked diffusion with bidirectional attention to text and images in one vocabulary.
In driving, masked diffusion decodes the plan: ReflectDrive~\cite{li2025reflectdrive} generates trajectory tokens from an action codebook and inpaints unsafe ones, ReflectDrive-2~\cite{wang2026reflectdrive2} learns to edit its own trajectory tokens with reinforcement learning, dVLM-AD~\cite{ma2025dvlmad} decodes structured reasoning and the trajectory together, Fast-dDrive~\cite{zhang2026fastddrive} decodes them in causally ordered blocks for speed, and WAM-Diff~\cite{xu2025wamdiff} adds a mixture of experts and policy optimization.
ReflectDrive, dVLM-AD and WAM-Diff build on LLaDA-V, and Fast-dDrive converts an autoregressive vision-language model to block diffusion; all of these models read the camera through a continuous vision encoder, can describe an image but not generate one, and do not model the future.
Discrete-WAM~\cite{yao2026discretewam} tokenizes both images and actions and edits action tokens in parallel, but generates future frames frame by frame.
Copilot4D~\cite{zhang2024copilot4d} learns a discrete diffusion world model, but on LiDAR tokens, and forecasts future point clouds by unmasking them rather than planning from the masked state.
All of them use masked diffusion as a decoder that unmasks tokens over several steps.
We use MMaDA-Parallel-M because bidirectional masked diffusion supports joint prediction of multiple future frames. The planner reads their masked states in one pass rather than iteratively unmasking them.

\section{Method}
\label{sec:method}

\begin{figure*}[t]
  \centering
  \includegraphics[width=\linewidth]{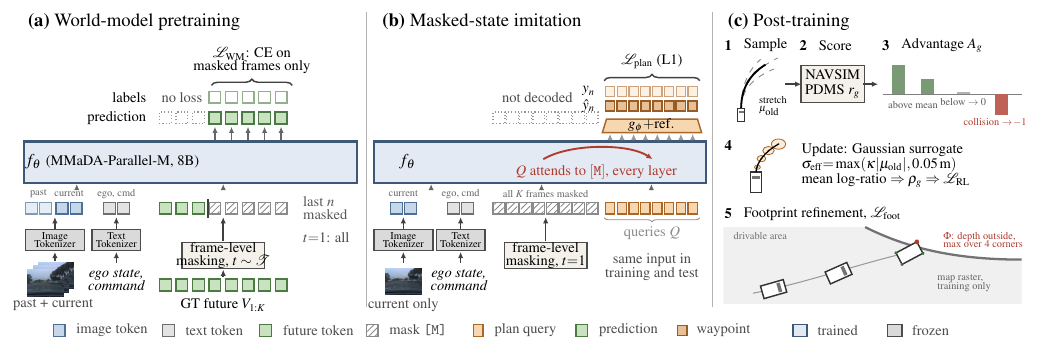}
  \caption{Training of \method. (a)~Frame-level masking replaces the last $n$ future frames with \mtok{} (all of them at $t=1$), and $\mathcal{L}_{\mathrm{WM}}$ is computed only there. (b)~Masked-state imitation uses the same input as at test time; $g_\phi$ predicts residuals, then the reference is added. (c)~Post-training with simulator rewards, then a footprint loss.}
  \label{fig:overview}
\end{figure*}

We build on MMaDA-Parallel-M~\cite{tian2025mmadaparallel}, a masked diffusion model over text and image tokens (\cref{sec:prelim}). Training comprises frame-level world-model pretraining (\cref{sec:wm}), masked-state imitation (\cref{sec:plan}), and simulator-reward post-training (\cref{sec:rl}); \cref{fig:overview,fig:infer} show training and inference.

\subsection{Preliminaries: Masked Diffusion}
\label{sec:prelim}

A masked diffusion model~\cite{nie2025llada} corrupts a clean token sequence $x_0$ by replacing each token with \mtok{} independently with probability $t\in[0,1]$,
\begin{equation}
  q_t(x_t^i\mid x_0^i) = (1-t)\,\mathbf{1}[x_t^i=x_0^i] + t\,\mathbf{1}[x_t^i=\mtok],
  \label{eq:fwd}
\end{equation}
and trains a bidirectional Transformer $p_\theta$ to recover the masked tokens from all the others,
\begin{equation}
  \mathcal{L}_{\mathrm{MDM}} = -\,\mathbb{E}_{t,x_t}\Big[\frac{1}{t}\sum_{i:\,x_t^i=\mtok}\log p_\theta(x_0^i\mid x_t)\Big].
  \label{eq:mdm}
\end{equation}
Generation starts from a fully masked sequence and runs several steps; each step predicts all masked tokens in parallel and keeps the most confident ones.
MMaDA~\cite{yang2025mmada} trains this objective on text and image tokens in a shared vocabulary.
Because $t$ can reach $1$, a fully masked segment is already a valid input for this model. Independent masking makes such inputs rare, so \cref{sec:wm} trains on them deliberately.

\subsection{Frame-Level Masked World Modeling}
\label{sec:wm}

Next-frame prediction~\cite{zhao2025pwm} can favor copying the most recent frame when adjacent observations are visually similar, rather than learning how the scene evolves. We therefore apply the backbone's masked diffusion objective to a suffix of whole future frames.

\paragraph{Tokenization and context.}
The frozen PWM tokenizer~\cite{zhao2025pwm} produces high-resolution anchor codes and compact dynamic codes. The anchor codes share MMaDA's existing 8,192-entry image codebook and enter its image-token range through an index offset. The dynamic codes use a separate 8,192-entry codebook: we append 8,192 rows to both the input embedding and output projection, initialized by copying the corresponding existing image-code rows.
The pretraining context is
\begin{equation}
  C_{\mathrm{WM}} = \big[\,I;\ s;\ c;\ \underbrace{V_{-P},\dots,V_{-1}}_{\text{past frames}},\ V_0\,\big],
  \label{eq:ctx}
\end{equation}
where $I$ contains the current frame's anchor tokens, $s$ and $c$ encode the ego state and navigation command, and $V_k=(v_k^1,\dots,v_k^L)$ is a compact grid of discrete dynamic-code indices for frame $k$. Here $k=0$ is current, $k<0$ is past, and $k=1,\dots,K$ indexes the future.
The anchor tokens $I$ provide current visual context; $V_0$ represents the same frame in the compact format predicted for future frames. Compact grids represent scene layout and motion while keeping all $K$ future frames within one pass. Ego state and navigation condition the forecast on the vehicle's motion and intended route; planning omits the past frames (\cref{sec:plan}).

\paragraph{Frame-level masking.}
We append the $K$ future grids $V_1,\dots,V_K$ to the context and replace the last $n=\max\big(1,\mathrm{round}(tK)\big)$ of them with mask tokens, for a ratio $t\sim\mathcal{T}$ (\cref{fig:overview}a):
\begin{equation}
  \tilde{x}_t = \big[\,C_{\mathrm{WM}};\ \underbrace{V_1,\dots,V_{K-n}}_{\text{visible future}};\ \underbrace{\mtok^{L},\dots,\mtok^{L}}_{n\ \text{masked frames}}\,\big],
  \label{eq:fwdframe}
\end{equation}
where $\mtok^{L}$ is a frame of $L$ mask tokens.
The loss is computed only on the masked frames:
\begin{equation}
  \mathcal{L}_{\mathrm{WM}} = -\,\mathbb{E}_{t\sim\mathcal{T}}\sum_{k=K-n+1}^{K}\log p_\theta\big(V_k \mid \tilde{x}_t\big).
  \label{eq:wm}
\end{equation}
Here $\log p_\theta(V_k\mid\tilde{x}_t)$ sums token-wise log-probabilities over frame $k$'s $L$ positions.
The distribution $\mathcal{T}$ sets $t=1$ with probability $\rho$ and otherwise follows the cosine schedule of MaskGIT~\cite{chang2022maskgit}, $t=\cos(\pi u/2)$ with $u\sim\mathcal{U}(0,1)$.
Partly masked inputs train the model to predict a future suffix conditioned on the observed context and a visible future prefix. Setting $t=1$ masks every future frame, matching the future-token configuration used for planning; $\rho$ controls how often this case is sampled.
Masking a contiguous future suffix removes access to later frames, requiring prediction beyond the visible sequence.
At $t=1$, every future frame is predicted directly from the context, $p_\theta(V_k\mid C_{\mathrm{WM}})$.
The objective needs no trajectory labels: the target is the video itself, and the ego state and the command are already recorded in driving logs; for Internet videos without them, these tokens are left out.

\begin{figure}[t]
  \centering
  \includegraphics[width=\linewidth]{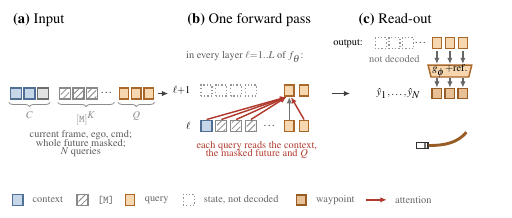}
  \caption{Inference in one forward pass: in every layer, each query attends to the context, the masked future and the other queries; the masked positions are never decoded.}
  \label{fig:infer}
\end{figure}

\subsection{Planning in the Masked State}
\label{sec:plan}

Planning appends one learned query per waypoint to the fully masked future and predicts a trajectory in one pass (\cref{fig:overview}b, \cref{fig:infer}).
The future is neither decoded before planning~\cite{zhao2025pwm} nor omitted from the planner input~\cite{li2025law,li2026drivevlaw0,zhao2026simwam}.

\paragraph{Planner input.}
The planner input consists of the context, the $K$ future frames, all masked, and $N$ learned query embeddings $Q=(q_1,\dots,q_N)$, one per waypoint:
\begin{equation}
  x_{\mathrm{plan}} = \big[C;\ \underbrace{\mtok^{L},\dots,\mtok^{L}}_{K\ \text{frames}};\ Q\big],
  \;
  C = \big[I;\ s;\ c;\ V_0\big].
  \label{eq:xplan}
\end{equation}
The numbers of future frames $K$ and waypoint queries $N$ need not be equal.
The future-token configuration matches \cref{eq:fwdframe} at $t=1$. Planning appends the queries and omits past frames from the context ($C$ instead of $C_{\mathrm{WM}}$).
Past frames can encourage extrapolation of the vehicle's recent motion at the expense of scene information~\cite{wen2020copycat,li2024egostatus}; the ego state still provides the current motion.
Removing past frames changes some relative token distances. Masked-state imitation trains the backbone, queries, and head on \cref{eq:xplan}, matching the input layout used at test time.

\paragraph{Reading the masked future.}
During world-model pretraining, let $h^{\mathrm{WM}}_{k,i}$ denote the hidden state at the $i$-th token of future frame $k$ when all future frames are masked. The output projection $W$ maps this state to a token distribution. At the population optimum of \cref{eq:wm}, assuming sufficient model capacity~\cite{ou2025absorbing},
\begin{equation}
  \mathrm{softmax}\big(W h^{\mathrm{WM}}_{k,i}\big)
  = p\big(v_k^i\mid C_{\mathrm{WM}}\big).
  \label{eq:marginal}
\end{equation}
During planning, the backbone instead processes $x_{\mathrm{plan}}$ in one pass, $H^{\mathrm{plan}}=f_\theta(x_{\mathrm{plan}})$. In each layer, queries attend to the context, masked-position states, and other queries (\cref{fig:infer}). The future inputs remain mask tokens, but their hidden states are updated using context. Queries read these states without decoding future tokens. Planning fine-tuning adapts the states to trajectory prediction; it need not preserve the equality in \cref{eq:marginal}.

\paragraph{Trajectory head.}
A small MLP $g_\phi$, shared by all queries, predicts waypoint offsets from a constant-speed straight-line reference~\cite{zheng2025resad}:
\begin{equation}
  \Delta\hat Y=g_\phi(H^{\mathrm{plan}}_Q),\qquad
  \hat Y=Y_{\mathrm{ref}}+\Delta\hat Y,
  \label{eq:plan}
\end{equation}
where $H^{\mathrm{plan}}_Q$ contains the last-layer query states. In ego coordinates, $y_{\mathrm{ref},n}=(vn\Delta t_{\mathrm{wp}},0)$, where $v$ is the current speed and $\Delta t_{\mathrm{wp}}=0.5\,\mathrm{s}$ is the waypoint interval. The reference is added before computing the imitation loss, so the head learns departures associated with braking and turning.
Each $\hat y_n\in\mathbb{R}^2$ contains only position. Heading is derived rather than regressed; for $n=2,\dots,N$,
\begin{equation}
  \hat\psi_n=\operatorname{atan2}\!\big(\hat y_{n}^{\,y}-\hat y_{n-1}^{\,y},\ \hat y_{n}^{\,x}-\hat y_{n-1}^{\,x}\big),
  \label{eq:heading}
\end{equation}
where superscripts $x$ and $y$ denote coordinate components, and $\hat\psi_1=\hat\psi_2$.

\paragraph{Masked-state imitation.}
We train the backbone, the queries and the head by imitation,
\begin{equation}
  \mathcal{L}_{\mathrm{plan}} = \sum_{n=1}^{N}\big\|\hat{y}_n-y_n\big\|_1,
  \label{eq:lplan}
\end{equation}
where $Y=(y_1,\dots,y_N)$ contains logged planar positions $y_n\in\mathbb{R}^2$.
The training input is \cref{eq:xplan}, the same as at test time.
We omit $\mathcal{L}_{\mathrm{WM}}$ during imitation. Gradients from the planning loss reach the masked states through the queries, allowing these representations to adapt to the planning objective (\cref{sec:abl-wm}).
Future-frame reconstruction also requires preserving visual appearance, whereas the planning loss supervises waypoint positions.

\subsection{Post-Training with Simulator Rewards}
\label{sec:rl}

Imitation learning minimizes trajectory error but does not directly optimize the evaluation metrics. We therefore post-train the planner using simulator rewards that account for collision avoidance, drivable-area compliance, and progress (\cref{fig:overview}c).

\paragraph{Trajectory sampling and a Gaussian surrogate.}
The head outputs one deterministic plan $\mu_\theta=\hat Y$. Following the trajectory-stretching approach of~\cite{zou2025diffusiondrivev2}, we sample candidates $\tau_g=\mu_{\mathrm{old}}\odot(1+\epsilon_g)$, where $\epsilon_g\sim\mathcal N(0,\kappa^2 I_2)$. The longitudinal and lateral perturbations are two scalars shared across every waypoint, preserving each path's smoothness. We use $\kappa=0.08$; the implementation floors this sampling scale at $0.04$.
For optimization, we use a coordinate-wise Gaussian surrogate with scales
$\sigma_{\mathrm{eff},d}=\max(\kappa|\mu_{\mathrm{old},d}|,0.05\,\mathrm m)$,
holding $\mu_{\mathrm{old}}$ and these scales fixed during each update.
The floor stabilizes density evaluation without adding noise to zero coordinates during sampling.
This surrogate ratio differs from the joint likelihood ratio of the correlated two-scalar sampler.
With $D=2N$, define $e_{g,d}=(\tau_{g,d}-\mu_{\mathrm{old},d})/\sigma_{\mathrm{eff},d}$ and $u_d=(\mu_{\theta,d}-\mu_{\mathrm{old},d})/\sigma_{\mathrm{eff},d}$. The GSPO-style ratio~\cite{zheng2025gspo} is
\begin{equation}
  \rho_g=\exp\!\left[\frac{1}{D}\sum_{d=1}^{D}
    \left(e_{g,d}u_d-\frac12u_d^2\right)\right],
  \label{eq:ratio}
\end{equation}
the geometric mean of per-coordinate density ratios. Only $(x,y)$ enters this surrogate; headings are derived from each sampled path. No sampling head is learned, and test-time prediction remains deterministic.

\paragraph{The simulator score as reward.}
We score each candidate with NAVSIM PDMS. For group reward mean $\bar r$ and standard deviation $s_r$, set $z_g=(r_g-\bar r)/(s_r+0.05)$ and
\begin{equation}
  A_g=\begin{cases}
    -1, & \mathrm{NC}_g<0.999,\\
    \max(z_g,0), & \text{otherwise}.
  \end{cases}
  \label{eq:advantage}
\end{equation}
Thus, at-fault collisions receive negative advantage, while other below-mean rewards are truncated to zero~\cite{zou2025diffusiondrivev2}. With $\tilde\rho_g=\operatorname{clip}(\rho_g,1-0.015,1+0.02)$, we minimize
\begin{equation}
  \mathcal{L}_{\mathrm{RL}}=-\frac{1}{G}\sum_{g=1}^{G}\min\!\big(\rho_gA_g,\ \tilde{\rho}_gA_g\big)+\lambda\,\|\mu_\theta-Y\|_1.
  \label{eq:rl}
\end{equation}
The last term anchors the plan to the logged trajectory, discouraging departures from it, while advantage truncation limits updates from below-mean non-collision samples.

\paragraph{Footprint refinement.}
PDMS checks whether the whole vehicle footprint stays in the drivable area, and the sampled plans give little signal for this. We therefore continue imitation briefly with a loss on the vehicle corners,
\begin{equation}
  \mathcal{L}_{\mathrm{foot}}=\frac{1}{|P|}\sum_{p\in P}\max_{c}\,\Phi\big(p+R(\psi_p)\,c\big),
  \label{eq:foot}
\end{equation}
where $P$ contains the waypoints and their midpoints, $c$ the four corners of the ego footprint, $\psi_p$ the heading between consecutive points, and $\Phi$ the distance outside the drivable area, read from a map raster used only in training.
We train with $\mathcal{L}_{\mathrm{plan}}+\beta\mathcal{L}_{\mathrm{foot}}$. Unlike the boundary loss of VAD~\cite{jiang2023vad} on the vehicle center, \cref{eq:foot} acts on the corners, where violations occur.

\section{Experiments}
\label{sec:exp}


We evaluate \method using non-reactive simulation on NAVSIM v1 and v2 and open-loop planning on nuScenes, with component ablations in \cref{sec:abl-wm}.

\begin{table*}[!ht]
\centering
\caption{Comparison on NAVSIM v1 \texttt{navtest}. NC: no at-fault collision, DAC: drivable area compliance, EP: ego progress, TTC: time to collision, C: comfort. Second group: methods that train a world model. Best in \textbf{bold}, excluding the human reference.}
\label{tab:navsim-v1}
\footnotesize
\renewcommand{\arraystretch}{0.92}
\setlength{\tabcolsep}{9pt}
\begin{tabular}{@{}llcccccc@{}}
\toprule
Method & Venue & NC$\uparrow$ & DAC$\uparrow$ & EP$\uparrow$ & TTC$\uparrow$ & C$\uparrow$ & PDMS$\uparrow$ \\
\midrule
\multicolumn{8}{@{}l}{\textit{Without world model}} \\
DiffusionDrive~\cite{liao2025diffusiondrive} & CVPR'25 & 98.2 & 96.2 & 82.2 & 94.7 & \textbf{100.0} & 88.1 \\
DiffusionDriveV2~\cite{zou2025diffusiondrivev2} & -- & 98.3 & 97.9 & 87.5 & 94.8 & 99.9 & 91.2 \\
ReCogDrive~\cite{li2026recogdrive} & ICLR'26 & 97.9 & 97.3 & 87.3 & 94.9 & \textbf{100.0} & 90.8 \\
R2SE~\cite{liu2026r2se} & TPAMI'26 & \textbf{99.0} & 97.9 & 86.8 & 96.4 & \textbf{100.0} & 91.6 \\
SparseDriveV2~\cite{sun2026sparsedrivev2} & ECCV'26 & 98.5 & 98.4 & 88.6 & 95.0 & 99.9 & \textbf{92.0} \\
\midrule
\multicolumn{8}{@{}l}{\textit{With world model}} \\
World4Drive~\cite{zheng2025world4drive} & ICCV'25 & 97.4 & 94.3 & 79.9 & 92.8 & \textbf{100.0} & 85.1 \\
PWM~\cite{zhao2025pwm} & NeurIPS'25 & 98.9 & 95.8 & 81.5 & 95.9 & \textbf{100.0} & 88.1 \\
WoTE~\cite{li2025wote} & ICCV'25 & 98.5 & 96.8 & 81.9 & 94.9 & 99.9 & 88.3 \\
SeerDrive~\cite{zhang2025seerdrive} & NeurIPS'25 & 98.8 & 98.6 & 84.2 & 95.8 & \textbf{100.0} & 90.7 \\
Epona~\cite{zhang2025epona} & ICCV'25 & 97.9 & 95.1 & 80.4 & 93.8 & 99.9 & 86.2 \\
DriveVLA-W0~\cite{li2026drivevlaw0} & ICLR'26 & 98.7 & \textbf{99.1} & 83.3 & 95.3 & 99.3 & 90.2 \\
ExploreVLA~\cite{sheng2026explorevla} & ECCV'26 & 98.8 & 98.4 & 83.5 & 96.5 & 99.9 & 90.4 \\
\midrule
\method (ours) & -- & 98.2 & 98.0 & \textbf{90.1} & \textbf{97.8} & 99.8 & \textbf{92.0} \\
\midrule
\textit{Human} & -- & 100.0 & 100.0 & 87.5 & 100.0 & 99.9 & 94.8 \\
\bottomrule
\end{tabular}
\end{table*}
\begin{table*}[!ht]
\centering
\caption{Comparison on NAVSIM v2 \texttt{navtest}. DDC: driving direction compliance, TLC: traffic light compliance, LK: lane keeping, HC: history comfort, EC: extended comfort. EPDMS$^*$\,/\,EPDMS: original\,/\,revised evaluation code (NAVSIM issue \#151); --: not reported.}
\label{tab:navsim-v2}
\footnotesize
\renewcommand{\arraystretch}{0.92}
\setlength{\tabcolsep}{4pt}
\begin{tabular}{@{}llccccccccccc@{}}
\toprule
Method & Venue & NC$\uparrow$ & DAC$\uparrow$ & DDC$\uparrow$ & TLC$\uparrow$ & EP$\uparrow$ & TTC$\uparrow$ & LK$\uparrow$ & HC$\uparrow$ & EC$\uparrow$ & EPDMS$^*\uparrow$ & EPDMS$\uparrow$ \\
\midrule
\multicolumn{13}{@{}l}{\textit{Without world model}} \\
ReCogDrive~\cite{li2026recogdrive} & ICLR'26 & 98.3 & 95.2 & 99.5 & 99.8 & 87.1 & 97.5 & 96.6 & \textbf{98.3} & 86.5 & 83.6 & -- \\
DriveSuprim~\cite{yao2026drivesuprim} & AAAI'26 & 97.8 & 97.9 & 99.5 & \textbf{99.9} & 90.6 & 97.1 & 96.6 & \textbf{98.3} & 77.9 & 86.0 & -- \\
DiffusionDriveV2~\cite{zou2025diffusiondrivev2} & -- & 97.7 & 96.6 & 99.2 & 99.8 & 88.9 & 97.2 & 96.0 & 97.8 & \textbf{91.0} & 85.5 & 87.5 \\
SparseDriveV2~\cite{sun2026sparsedrivev2} & ECCV'26 & 98.1 & 98.1 & \textbf{99.6} & 99.8 & \textbf{91.1} & 97.3 & 96.9 & 98.2 & 78.4 & \textbf{86.7} & \textbf{90.1} \\
\midrule
\multicolumn{13}{@{}l}{\textit{With world model}} \\
WAM-Flow~\cite{xu2026wamflow} & CVPR'26 & 98.5 & 94.5 & 99.5 & 99.8 & 86.9 & 96.8 & \textbf{97.4} & 97.6 & 73.9 & 84.7 & -- \\
Epona~\cite{zhang2025epona} & ICCV'25 & 97.1 & 95.7 & 99.3 & 99.7 & 88.6 & 96.3 & 97.0 & 98.0 & 67.8 & -- & 85.1 \\
DriveVLA-W0~\cite{li2026drivevlaw0} & ICLR'26 & 98.5 & \textbf{99.1} & 98.0 & 99.7 & 86.4 & \textbf{98.1} & 93.2 & 97.9 & 58.9 & 86.1 & -- \\
\midrule
\method (ours) & -- & \textbf{98.6} & 98.0 & 99.4 & 99.8 & 87.1 & 98.0 & 97.2 & 98.1 & 80.0 & -- & 89.3 \\
\midrule
\textit{Human} & -- & 100.0 & 100.0 & 99.8 & 100.0 & 87.4 & 100.0 & 100.0 & 98.1 & 90.1 & 90.3 & -- \\
\bottomrule
\end{tabular}
\end{table*}
\subsection{Experimental Setup}
\label{sec:setup}

\paragraph{Datasets and metrics.}
We pretrain on 1.34M clips from OpenDV-YouTube~\cite{yang2024genad} and 692k clips from nuPlan~\cite{caesar2021nuplan}, without trajectory labels, and fine-tune planners on NAVSIM~\cite{dauner2024navsim}, a non-reactive pseudo-simulation built on nuPlan, and on nuScenes, using the front camera only.
We train on \texttt{navtrain} (103k scenes) and report on \texttt{navtest} (12{,}146 scenes): the PDM score (PDMS) for NAVSIM v1, which combines no at-fault collision (NC), drivable-area compliance (DAC), time-to-collision (TTC), comfort (C) and ego progress (EP), and the extended score (EPDMS) for v2.
On nuScenes we report L2 error and collision rate at 1, 2 and 3\,s under the ST-P3~\cite{hu2022stp3}/VAD~\cite{jiang2023vad} protocol that PWM~\cite{zhao2025pwm} follows, with the same collision implementation, so \cref{tab:nusc} follows one protocol; we exclude numbers under the UniAD~\cite{hu2023uniad} protocol, which are about twice as large for the same model~\cite{guo2024uad}.

\paragraph{Implementation details.}
The backbone is MMaDA-Parallel-M~\cite{tian2025mmadaparallel} (8B, fully bidirectional), the frozen PWM tokenizer~\cite{zhao2025pwm} encodes the current frame into 448 tokens and every frame into a compact grid of $L=28$, and the planning queries and trajectory head add about 25M parameters, separate from the dynamic-code vocabulary expansion (\cref{sec:wm}).
All runs use 8 B200 GPUs with an effective batch of 32.
In pretraining (learning rate $10^{-4}$), OpenDV clips carry no ego state or command, so those tokens are dropped and $\rho=\tfrac{1}{2}$; the nuPlan stage then uses $P=8$ past and $K=8$ future frames at 0.5\,s intervals for 75k steps at $\rho=1$. Its future horizon matches NAVSIM planning; imitation and inference omit the past frames. The nuPlan pretraining clips come from the OpenScene \texttt{trainval} logs, so no \texttt{navtest} scene is seen.
For NAVSIM we then run masked-state imitation on \texttt{navtrain} ($N=8$ waypoints over 4\,s, 10 epochs, learning rate $10^{-5}$), post-training on 25k scenes ($G=8$, $\kappa=0.08$, $\lambda=0.1$, learning rate $5\times10^{-6}$) and one epoch of footprint refinement ($\beta=8$).
For nuScenes we follow PWM with imitation only ($N=6$ waypoints over 3\,s).

\subsection{Main Results}
\label{sec:main}


\begin{table*}[t]
\centering
\caption{Open-loop planning on nuScenes \texttt{val}. WM: trains a world model. Ego: ego status as input.}
\label{tab:nusc}
\footnotesize
\renewcommand{\arraystretch}{0.92}
\setlength{\tabcolsep}{4pt}
\begin{tabular}{@{}llcccccccccc@{}}
\toprule
& & & & \multicolumn{4}{c}{L2 (m)$\downarrow$} & \multicolumn{4}{c}{Collision (\%)$\downarrow$} \\
\cmidrule(lr){5-8}\cmidrule(lr){9-12}
Method & Venue & WM & Ego & 1s & 2s & 3s & Avg. & 1s & 2s & 3s & Avg. \\
\midrule
Drive-OccWorld~\cite{yang2025driveoccworld} & AAAI'25 & \checkmark &  & 0.25 & 0.44 & 0.72 & 0.47 & 0.03 & 0.08 & 0.22 & 0.11 \\
LAW~\cite{li2025law} & ICLR'25 & \checkmark &  & 0.24 & 0.46 & 0.76 & 0.49 & 0.08 & 0.10 & 0.39 & 0.19 \\
Impromptu VLA~\cite{chi2025impromptu} & NeurIPS'25 &  & \checkmark & \textbf{0.13} & \textbf{0.27} & 0.53 & \textbf{0.30} & -- & -- & -- & -- \\
PWM~\cite{zhao2025pwm} & NeurIPS'25 & \checkmark & \checkmark & 0.20 & 0.38 & 0.65 & 0.41 & \textbf{0.01} & \textbf{0.02} & \textbf{0.09} & \textbf{0.04} \\
VGGDrive~\cite{wang2026vggdrive} & CVPR'26 &  & \checkmark & 0.14 & 0.28 & 0.51 & 0.31 & 0.02 & 0.10 & 0.55 & 0.22 \\
ColaVLA~\cite{peng2026colavla} & CVPR'26 &  & \checkmark & 0.14 & \textbf{0.27} & 0.50 & \textbf{0.30} & 0.04 & 0.17 & 0.47 & 0.23 \\
\midrule
\method (ours) & -- & \checkmark & \checkmark & 0.17 & 0.29 & \textbf{0.47} & 0.31 & 0.02 & 0.05 & 0.10 & 0.06 \\
\bottomrule
\end{tabular}
\end{table*}

\paragraph{NAVSIM v1.}
\cref{tab:navsim-v1} compares \method with published planners.
With a single front camera and one forward pass, \method reaches 92.0 PDMS, the highest among the world-model planners compared in \cref{tab:navsim-v1}.
It is 3.9 points above PWM (88.1), which uses the same tokenizer and camera, and SparseDriveV2 reaches the same score with three cameras and no world model.
\method has the highest EP and TTC, while its DAC and NC are below the best methods.

\paragraph{NAVSIM v2.}
\cref{tab:navsim-v2} lists each method under the release of the evaluation code it was measured with.
Under the corrected release, \method reaches 89.3 EPDMS, above DiffusionDriveV2 (87.5) and Epona (85.1) and below SparseDriveV2 (90.1, three cameras).
Its extended comfort (80.0) is higher than that of the other world-model methods but below DiffusionDriveV2 (91.0).

\paragraph{nuScenes.}
On nuScenes (\cref{tab:nusc}), \method reaches 0.31\,m average L2 from a single camera, the lowest among world-model planners (PWM 0.41) and comparable to recent VLM-based planners (VGGDrive 0.31, ColaVLA 0.30), with the same continuous head as on NAVSIM. PWM keeps the lowest collision rate (0.04\%) and \method is second (0.06\%).

\begin{figure}[t]
  \centering
  \includegraphics[width=\linewidth]{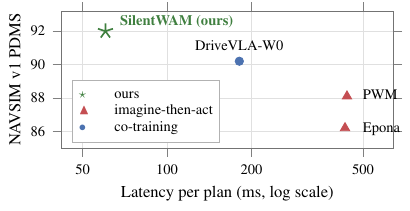}
  \caption{Accuracy against latency on NAVSIM v1 \texttt{navtest}. Latency is end-to-end per plan on one B200 GPU (bf16, batch size 1). \method achieves the highest PDMS and lowest latency among the compared world-model planners.}
  \label{fig:latency}
\end{figure}

\paragraph{Efficiency.}
\method plans in 60.3\,ms end-to-end on one B200 GPU (bf16, batch size 1), from the camera image to the trajectory, including the image tokenizer; the planner itself is a single forward pass. The backbone has 8B parameters, with vocabulary expansion described in \cref{sec:wm}.
\cref{fig:latency} compares this with other world-model planners under the same protocol.
Imagine-then-act planners are slower, as they generate the future before planning: PWM takes 438\,ms and Epona 431\,ms.
DriveVLA-W0 plans in one pass but on a larger action decoder, at 181\,ms.

\subsection{Qualitative Results}

\begin{figure*}[t]
\centering
\includegraphics[width=\linewidth]{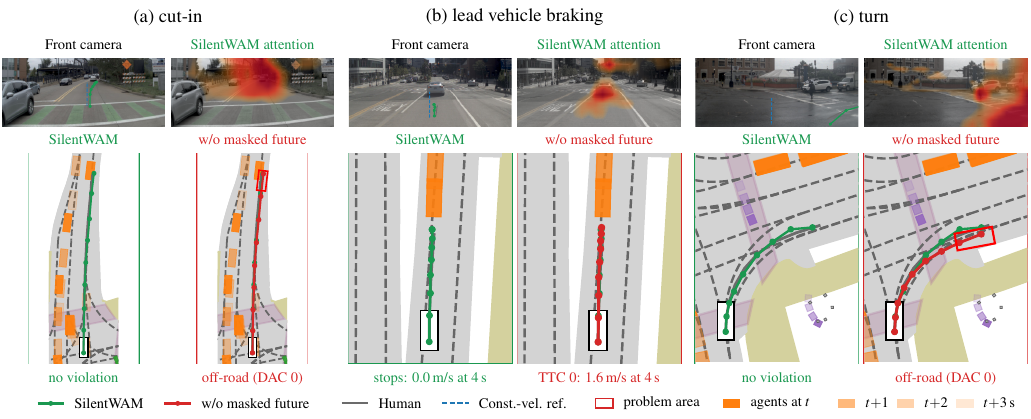}
\caption{Qualitative results on NAVSIM \texttt{navtest}: \method vs.\ the model without the masked future (\cref{tab:train-wm}).}
\label{fig:qual}
\end{figure*}

\cref{fig:qual} compares \method with the model trained without the masked future on three \texttt{navtest} scenes. In (a), the road narrows and a vehicle from the left lane merges into the ego lane within 4\,s; in (c), the ego vehicle turns right in the rain while a pedestrian crosses. Without the masked future, the plan leaves the drivable area, about 43\,m ahead in (a) and at the exit of the turn in (c); the plans of \method stay on the road. In (b), the baseline still moves at 1.6\,m/s after 4\,s behind the braking lead vehicle, whereas \method stops. The planning queries attend to the road ahead in (a), the lead vehicle in (b), and the pedestrian and the exit lane in (c).

\subsection{Ablation Studies}
\label{sec:abl-wm}

\begin{table}[t]
\centering
\caption{Ablation of the main components. PT\,/\,IL: $\mathcal{L}_{\mathrm{WM}}$ in pretraining\,/\,masked-state imitation. \mtok$^{K}$: masked future frames in the planner input. $\hat{V}$: future decoded before planning. Post.: post-training (\cref{sec:rl}; NAVSIM only). nuScenes: average L2 (m) and collision rate (\%). Latency per plan on one B200 GPU.}
\label{tab:train-wm}
\footnotesize
\renewcommand{\arraystretch}{0.92}
\setlength{\tabcolsep}{4.5pt}
\begin{tabular}{@{}cccccc|c|cc|c@{}}
\toprule
& \multicolumn{2}{c}{$\mathcal{L}_{\mathrm{WM}}$} & & & & NAVSIM & \multicolumn{2}{c|}{nuScenes} & \\
\cmidrule(lr){2-3}
& PT & IL & \mtok$^{K}$ & $\hat{V}$ & Post. & PDMS$\uparrow$ & L2$\downarrow$ & Col.$\downarrow$ & ms$\downarrow$ \\
\midrule
(a) & $\times$ & $\times$ & \checkmark & $\times$ & $\times$ & 83.6 & 0.55 & 0.51 & 60.3 \\
(b) & \checkmark & $\times$ & $\times$ & $\times$ & $\times$ & 90.9 & 0.41 & 0.18 & 47.5 \\
(c) & \checkmark & $\times$ & \checkmark & \checkmark & $\times$ & 91.3 & 0.32 & 0.10 & -- \\
(d) & \checkmark & \checkmark & \checkmark & $\times$ & $\times$ & 90.8 & 0.35 & 0.08 & 60.3 \\
(e) & \checkmark & $\times$ & \checkmark & $\times$ & $\times$ & 91.1 & \textbf{0.31} & \textbf{0.06} & 60.3 \\
(f) & \checkmark & $\times$ & \checkmark & $\times$ & \checkmark & \textbf{92.0} & -- & -- & 60.3 \\
\bottomrule
\end{tabular}
\end{table}

Ablations report NAVSIM v1 PDMS and nuScenes L2 and collision rate, without post-training unless marked.

\paragraph{Main components.}
\cref{tab:train-wm} ablates each component.
\begin{itemize}\itemsep1pt \parskip0pt \topsep2pt
\item \textbf{World-model pretraining improves planning} (a vs.\ e). Without it, PDMS falls from 91.1 to 83.6 and the nuScenes collision rate rises from 0.06\% to 0.51\%. Pretraining also adapts the backbone to the compact frame tokens.
\item \textbf{Masked future states reduce nuScenes collisions} (b vs.\ e). Dropping it from the planner input (the co-training setting on our backbone) barely changes NAVSIM PDMS (90.9 vs.\ 91.1), but raises the nuScenes collision rate to 0.18\% while reducing latency by 12.8\,ms.
\item \textbf{Future decoding offers limited additional benefit} (c vs.\ e). Decoding adds 8 unmasking steps and improves PDMS only from 91.1 to 91.3, while increasing nuScenes L2 and collision rate.
\item \textbf{Retaining the world-model loss slightly reduces performance} (d vs.\ e). Keeping $\mathcal{L}_{\mathrm{WM}}$ lowers PDMS from 91.1 to 90.8. We therefore optimize the masked representations using only the planning loss during imitation (\cref{sec:plan}).
\item \textbf{Post-training helps} (e vs.\ f). It adds 0.9 PDMS (91.1 to 92.0) at the same latency.
\end{itemize}

\paragraph{Temporal setting of the future.}
\cref{tab:abl}(a) varies the number $K$ and the interval $\Delta t$ of the future frames, in both pretraining and planning.
For NAVSIM, the default $K{=}8$ frames at 0.5\,s intervals align with the eight waypoint timestamps.
With a 2\,s future horizon, the masked frames no longer cover the full planned trajectory, and the nuScenes collision rate rises from 0.06\% to 0.15\%.
With the 4\,s horizon kept but a 1.0\,s interval, the collision rate stays low (0.09\%), but PDMS drops to 90.0; this setting provides only one future frame within the first second.
An 8\,s horizon at the same $K$ is worse on both benchmarks (90.6 PDMS; 0.36\,m and 0.11\%), with some future frames beyond the planned trajectory.

\paragraph{Where the gains come from.}\label{sec:analysis}\label{sec:abl-plan}
We break the nuScenes L2 error down by maneuver (\cref{tab:abl}(b)).
World-model pretraining helps in every case, but far more on turns than going straight: it reduces the error from 1.80 to 0.40\,m on left turns and from 1.70 to 0.35\,m on right turns, against 0.34 to 0.27\,m on straight segments. The larger gains on turns suggest that pretraining is especially useful for trajectories that depart from the constant-speed reference.

\begin{table}[t]
\centering
\caption{Ablations on the masked future. \textbf{(a)} Number $K$ and spacing $\Delta t$ of the future frames (horizon $K\Delta t$); each row pretrains its own world model, imitation learning only. \textbf{(b)} nuScenes L2 error (m) by maneuver, with the share of clips, without and with world-model pretraining; measured separately from the overall L2 in \cref{tab:train-wm}.}
\label{tab:abl}
\footnotesize
\textbf{(a) Future frames}\\[3pt]
\renewcommand{\arraystretch}{0.92}
\setlength{\tabcolsep}{4pt}
\begin{tabular}{@{}ccc|c|cc|c@{}}
\toprule
& & & NAVSIM & \multicolumn{2}{c|}{nuScenes} & \\
$K$ & $\Delta t$ & Horizon & PDMS$\uparrow$ & L2$\downarrow$ & Col.$\downarrow$ & ms$\downarrow$ \\
\midrule
4 & 0.5\,s & 2\,s & 90.7 & 0.34 & 0.15 & 54.9 \\
4 & 1.0\,s & 4\,s & 90.0 & 0.32 & 0.09 & 55.4 \\
8 & 0.5\,s & 4\,s & \textbf{91.1} & \textbf{0.31} & \textbf{0.06} & 60.3 \\
8 & 1.0\,s & 8\,s & 90.6 & 0.36 & 0.11 & 60.6 \\
\bottomrule
\end{tabular}

\vspace{6pt}
\textbf{(b) Per-maneuver L2}\\[3pt]
\renewcommand{\arraystretch}{0.95}
\setlength{\tabcolsep}{8pt}
\begin{tabular}{@{}lccc@{}}
\toprule
Maneuver & Clips & w/o pretrain & w/ pretrain \\
\midrule
Straight & 82.4\% & 0.34 & 0.27 \\
Left turn & 7.3\% & 1.80 & \textbf{0.40} \\
Right turn & 10.3\% & 1.70 & \textbf{0.35} \\
\bottomrule
\end{tabular}
\end{table}

\section{Conclusion}
\label{sec:conclusion}

\method plans from the masked state of a masked diffusion world model: frame-level masked world modeling makes the fully masked future a regular training input, and learned queries read it in the same forward pass. In ablations, retaining the masked future lowers the nuScenes collision rate from 0.18\% to 0.06\%. Explicit future decoding adds generation steps for only a marginal gain in NAVSIM PDMS and increases the nuScenes collision rate to 0.10\%.

\paragraph{Limitations.}
\method uses a single front camera and a frozen tokenizer, leaving surround-view and LiDAR to future work. Post-training explores only two-axis rescalings of the current plan. Evaluation uses non-reactive NAVSIM simulation and open-loop nuScenes metrics; reactive closed-loop driving is untested.

{
    \small
    \bibliographystyle{ieeenat_fullname}
    \bibliography{main}
}


\end{document}